\documentclass{article}
\usepackage{spconf,amsmath,amssymb,graphicx,booktabs,multirow}

\title{LiveProBench: Can Streaming Video Models Really Interact Like Humans?}

\name{Kaixuan Du$^{1,2,*,\dagger}$, Xin Wan$^{2,*,\ddagger}$, Hang Zhang$^{1}$,
Meng Cao$^{1}$, Dai Guan$^{2}$, Ming Chen$^{2}$, YuKun Wang$^{1,\ddagger}$
\thanks{$^*$Equal contribution.}
\thanks{$^\dagger$Work done during an internship at Alibaba.}
\thanks{$^\ddagger$Corresponding author.}
}
\address{
  $^1$School of Automation Science and Electrical Engineering, Beihang University \\
  $^2$Qwen Business Unit of Alibaba}

\begin{document}
\maketitle

\begin{abstract}
Streaming video understanding requires models to process continuous multimodal input while maintaining temporal context. Existing evaluations are predominantly reactive: they query a model at a selected timestamp and therefore do not assess when it should respond. Proactive interaction instead requires monitoring a standing request, responding within an appropriate interval after the target event, and otherwise remaining silent. We introduce LiveProBench, which evaluates models at one-second stream intervals without an explicit response cue. Its six subtasks vary trigger ambiguity and timing tolerance. Event Sensitivity geometrically combines response and silence rates on the same recording; four window-based subtasks distinguish early, in-window, and missed responses; and Duplicate Counting penalizes omissions and repetitions. Premature responses outnumber missed responses for half of the evaluated models, revealing a substantial gap in the temporal decision-making required for human-like interaction.
\end{abstract}


\section{Introduction}
\label{sec:intro}

Streaming video models must update predictions from incoming multimodal evidence and prior context. Most benchmarks, however, advance the stream to an evaluator-selected timestamp before posing a query, thereby assessing partial-stream question answering rather than whether a response is warranted.

\begin{figure}[t]
\centering
\includegraphics[width=\columnwidth]{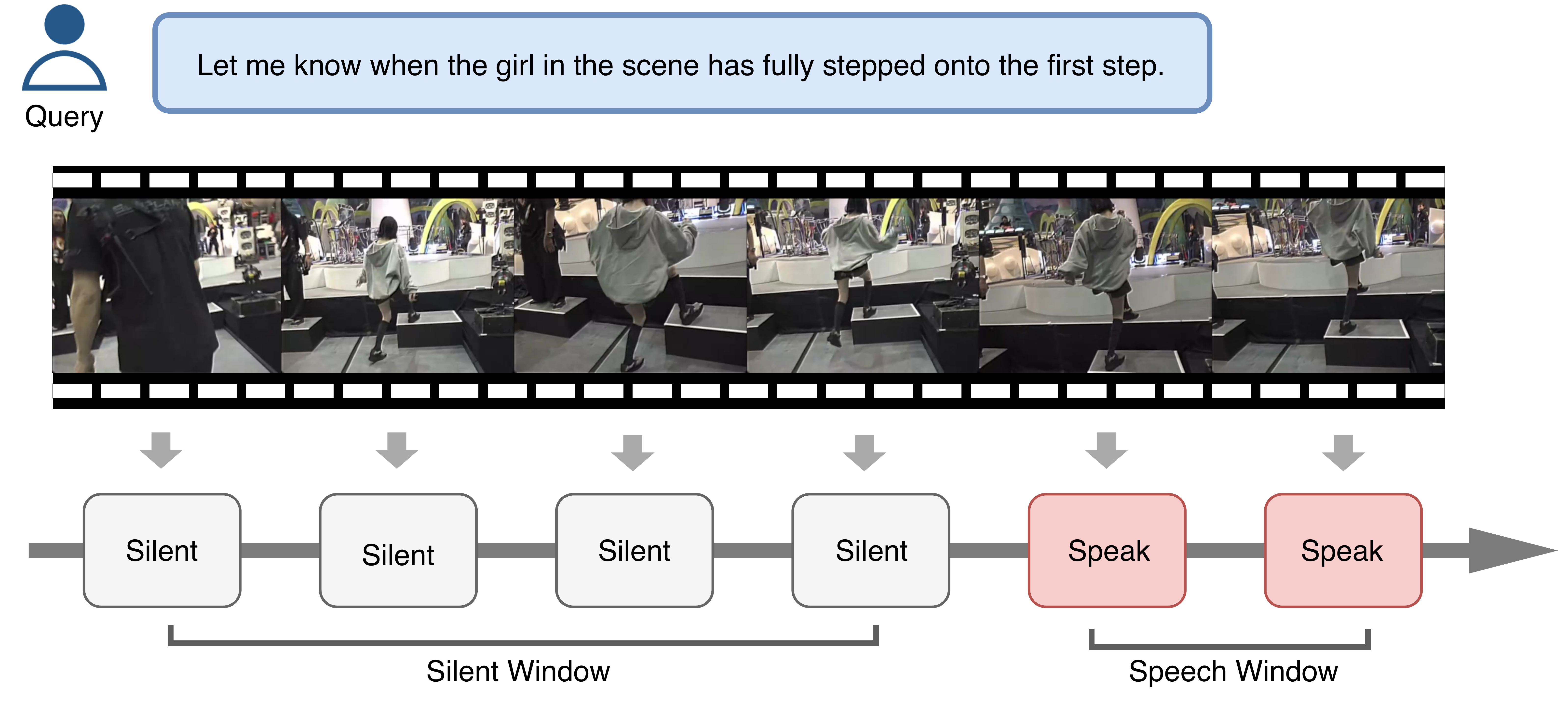}
\caption{Proactive streaming interaction under a standing request. The model is evaluated at successive stream updates and should remain silent until the requested state becomes observable, then respond within the annotated speech window.}
\label{fig:protocol}
\end{figure}

Standing requests make this distinction explicit. After receiving a request such as reporting when a kettle boils, a model must remain silent until sufficient evidence appears and then respond within an acceptable interval. Proactive interaction is therefore a temporal decision problem with complementary premature and missed-response errors; responding continuously incurs false positives. Figure~\ref{fig:protocol} illustrates this full-sequence evaluation.

Recent methods have enabled online video processing with bounded computational cost. VideoLLM-online \cite{chen24}, Flash-VStream \cite{zhang24}, and StreamForest \cite{zeng25} process video incrementally and answer queries from partial streams. Recent studies also examine response timing. StreamPro formulates this problem as decision-making under partial observations, balancing early prediction against the accumulation of sufficient evidence \cite{li26}. Related approaches investigate duet-style interaction \cite{wang24videollm}, separate perception, decision, and reaction modules \cite{qian25}, readiness-aware answering \cite{azad26}, and adaptation of offline models for proactive interaction \cite{wang25}. Several systems further define model-specific silence tokens \cite{lu26,wang26mossvl,yao26joyai}. Evaluation of these systems must distinguish response quality from the policy governing whether and when a response is generated.

Existing protocols do not fully separate these factors. Streaming benchmarks \cite{li25,lin24,huang25ovbench} specify the query timestamp, excluding preceding silence and premature responses from evaluation. Proactive suites \cite{zhang26streamarena} assess self-initiated responses, and OmniMMI \cite{wang25omnimmi} introduces the proactive alerting setting. However, response and silence conditions are typically evaluated on different recordings, which can confound response policy with video content. Low-frequency response policies may also obtain moderate average scores despite inaccurate timing. These limitations motivate paired, time-resolved measurement of premature, timely, and missed responses.

LiveProBench evaluates every response opportunity at one-second resolution. Its contributions are:

\begin{itemize}
\itemsep0pt
\item A per-second evaluation protocol without explicit response cues that preserves each model's native silence mechanism.
\item Six subtasks that vary response-time tolerance and trigger clarity, from events localized to one second to extended or ambiguous activities.
\item A paired subtask that geometrically aggregates response and silence rates on the same recording; four window-based subtasks penalize premature and missed responses, while DC penalizes omissions and repetitions.
\item A comparative evaluation identifying premature responses as the predominant error and quantifying the effect of paired scoring on model rankings.
\end{itemize}

\section{The LiveProBench suite}
\label{sec:bench}

\subsection{Protocol}
\label{ssec:protocol}

Each test instance consists of a continuous stream paired with a standing request. The stream is presented in one-second intervals, and the model is evaluated after each interval without being told whether the target event has occurred. It must therefore decide at every step whether to respond or remain silent. The initial interval establishes the standing request and is excluded because it does not provide a comparable response decision. Reference labels are assigned independently at each interval and remain unchanged by earlier model outputs; the conversion of these decisions into example-level outcomes is defined in Section~\ref{ssec:metrics}. The model receives at most the preceding 30 seconds of context and providing a common temporal horizon.

\subsection{Two axes, six subtasks}
\label{ssec:axes}

Table~\ref{tab:suite} summarizes these dimensions. \emph{Tolerance} allows zero delay, a fixed three-second delay, or a delay determined by the event duration. \emph{Semantic clarity} characterizes how precisely the trigger can be localized. It ranges from events with a single decisive second, such as contact, impact, or motion completion, through activities annotated as \emph{underway}, to anomaly requests that use the same generic sentence across all examples.

For ambiguous triggers, the response window follows event duration, accommodating onset uncertainty but conflating it with model delay. OR instead applies three annotation criteria to identify the earliest second at which the requested event becomes unambiguously observable, followed by a fixed three-second window. A response one second before onset is \emph{early}, whereas a response in the fourth second is \emph{missed}. Higher performance on OR than on the zero-tolerance subtask may indicate a latency limitation, although the comparison is indirect because the subtasks use disjoint recordings.

We refer to the six subtasks by the codes Event Sensitivity (ES), Onset Response (OR), Window Response (WR), Anomaly Alert (AA), Sustained Silence (SS), and Duplicate Counting (DC). Each contributes exactly one score to the reported average, and the abbreviations are used throughout the paper.

\begin{table}[t]
\caption{The six subtasks. Each defines when a response is expected and how response delay is tolerated; semantic clarity ranges from a single decisive second to extended or ambiguous activities. The \emph{decisions} column gives the number of evaluated one-second intervals. Tolerance and window width are in seconds. $^\dagger$Response and silence are evaluated on the same recording.}
\label{tab:suite}
\centering
\footnotesize
\setlength{\tabcolsep}{2.5pt}
\begin{tabular*}{0.95\columnwidth}{@{\hspace{8pt}\extracolsep{\fill}}llrrll@{\hspace{8pt}}}
\toprule
subtask & target condition & examples & decisions & tol. & width \\
\midrule
ES & First evidence$^\dagger$ & 216 & 1{,}387 & none & 1 \\
OR & Event onset              & 218 & 1{,}992 & 3\,s   & 3 \\
WR & Response window          &  80 & 2{,}530 & event  & 6--25 \\
AA & Anomaly alert            & 238 & 1{,}977 & event  & 1--17 \\
SS & Extended silence         &  92 & 9{,}384 & event  & 4--22 \\
DC & Repeated occurrence      &  67 &    876  & n/a    & n/a \\
\bottomrule
\end{tabular*}
\end{table}

SS and DC target response behaviors not captured by the other subtasks. SS measures whether a model remains silent during an extended prefix in which the requested event has not occurred. Its response window opens after 60 to 194 consecutive seconds of verified silence, and each SS example contains 102 evaluated decisions on average, compared with 6 to 32 for the other subtasks. This design jointly evaluates sustained silence and subsequent event recognition: a model that always remains silent misses the eventual event, whereas a response during the prefix violates the silence condition. DC evaluates whether a model tracks repeated occurrences without over-reporting. Let $\mathcal{D}_{\mathrm{DC}}$ denote the DC evaluation set, and let $c_i$ and $\hat c_i$ be the reference and predicted occurrence counts for example $i$. We define example-level exact-match accuracy as
\begin{equation}
\mathrm{DC}
=\frac{1}{|\mathcal{D}_{\mathrm{DC}}|}
\sum_{i\in\mathcal{D}_{\mathrm{DC}}}
\mathbf{1}(\hat c_i=c_i).
\end{equation}
Thus, an example is correct only when its predicted total equals the reference count.

\subsection{Data}
\label{ssec:data}

For each example, the released annotations specify the source recording, temporally localized event labels, the standing request, and reference responses. Every example is associated with a distinct video, and videos are not shared across subtasks, preventing any recording from contributing to multiple scores. Stream durations range from 4 to 207 seconds, totaling 5.4 hours and 18{,}146 evaluated one-second decisions.

Requests are written for individual recordings and undergo independent review. A second annotator examines each recording without access to the proposed request or target time, and a prefix-level verification confirms that no earlier interval satisfies the request. Examples that fail this criterion are excluded. The anomaly subtask uses one request sentence throughout, reducing variation attributable to wording while retaining differences in video content, event timing, and model capability.

\subsection{Metrics}
\label{ssec:metrics}

Because a model that responds at every second satisfies every response-required condition, response rate alone is insufficient. In each window-based subtask, the first non-silent response determines the example-level outcome: a response before the annotated window is \emph{early}, one within the window is \emph{in-window}, and a response after the window or no response is \emph{missed}. The corresponding accuracy is the fraction of examples classified as in-window, while DC uses the exact-match accuracy defined above.

ES jointly evaluates responsiveness and restraint. Let $\mathcal{A}$ denote its set of annotated alerts and $\mathcal{G}$ the set of silence segments formed by the pre- and post-alert intervals. For alert $a$, let $r_a$ indicate a response at the annotated alert time; for segment $g$, let $q_g$ indicate that the model remains silent throughout the segment. We define
\begin{equation}
\begin{gathered}
  \begin{alignedat}{2}
    R_{\mathrm{ES}} &= \frac{1}{|\mathcal{A}|}\sum_{a\in\mathcal{A}}r_a, &\quad
    Q_{\mathrm{ES}} &= \frac{1}{|\mathcal{G}|}\sum_{g\in\mathcal{G}}q_g,
  \end{alignedat} \\
  S_{\mathrm{ES}} = \sqrt{R_{\mathrm{ES}}Q_{\mathrm{ES}}}.
\end{gathered}
\end{equation}
Here, $r_a,q_g\in\{0,1\}$. The geometric mean requires both components of ES to be high and therefore discourages degenerate policies that respond indiscriminately or remain uniformly silent. AVG is the arithmetic mean of $S_{\mathrm{ES}}$ and the five remaining subtask accuracies.

Pairing response and silence conditions on the same recording controls for content, wording, and recording difficulty; separate recordings would compare distinct sample populations. The four window-based subtasks penalize premature responses as \emph{early} and the absence of a timely response as \emph{missed}, whereas DC penalizes deviations from the reference occurrence count.

\section{Experiments}
\label{sec:exp}

\subsection{Setup}
\label{ssec:setup}

\textbf{Supervised fine-tuning.} Building on recent work in streaming video instruction tuning \cite{xia25}, we construct a training corpus of 196k sequences from public video-QA and video-caption datasets. Each sequence represents a continuous interaction in which the model either answers an explicit query or determines whether a proactive response is warranted. Silence decisions and explicit-query responses are supervised throughout, whereas proactive trajectories retain only the final non-silent response, avoiding repeated positive targets for the same event.

For sequence $i$, let $\mathcal{T}_i$ denote the supervised output-token indices and $\mathcal{S}_i\subseteq\mathcal{T}_i$ those associated with silent decisions. Given target token $y_{it}$ and multimodal history $h_{it}$, we use
\begin{equation}
\begin{split}
\mathcal{L}_{\mathrm{SFT}}
&=-\frac{1}{\sum_i |\mathcal{T}_i|}
\sum_i\sum_{t\in\mathcal{T}_i}w_{it}
\log p_\theta(y_{it}\mid h_{it}),\\
w_{it}
&=\begin{cases}
\max(1,|\mathcal{S}_i|)^{-1}, & t\in\mathcal{S}_i,\\
1, & \text{otherwise}.
\end{cases}
\end{split}
\label{eq:sft}
\end{equation}
This weighting normalizes the aggregate contribution of silent targets within each sequence and remains defined when a sequence contains no silent target. We evaluate M4 from OmniMMI \cite{wang25omnimmi}, ROMA \cite{tian26}, MOSS-VL-Realtime \cite{wang26mossvl}, AURA \cite{lu26}, JoyAI-VL-Interaction \cite{yao26joyai}, and our Qwen3-Omni-SFT. Our model uses Qwen3-Omni-30B-A3B as its base model and is optimized with Eq.~\eqref{eq:sft}. None of the evaluated models is further fine-tuned on LiveProBench.

Across models, we align visual content, frame order, input resolution, and decoding settings wherever their architectures permit. The primary comparison disables audio for all systems; Table~\ref{tab:main} additionally reports audio-enabled variants of ROMA and Qwen3-Omni-SFT. We retain each model's established prompting and silence criterion because replacing them would confound proactive behavior with adaptation to a new instruction format.

Two structural differences remain. M4 uses prompted silence because it has no silence token, so its silence scores are not directly comparable; treating ``nothing to report'' as silence would conflate a textual response with no response. ROMA uses a gate-head probability rather than a token, making its silence rate threshold-dependent.

\subsection{Results}
\label{ssec:results}

\begin{figure}[b]
\centering
\includegraphics[width=\columnwidth]{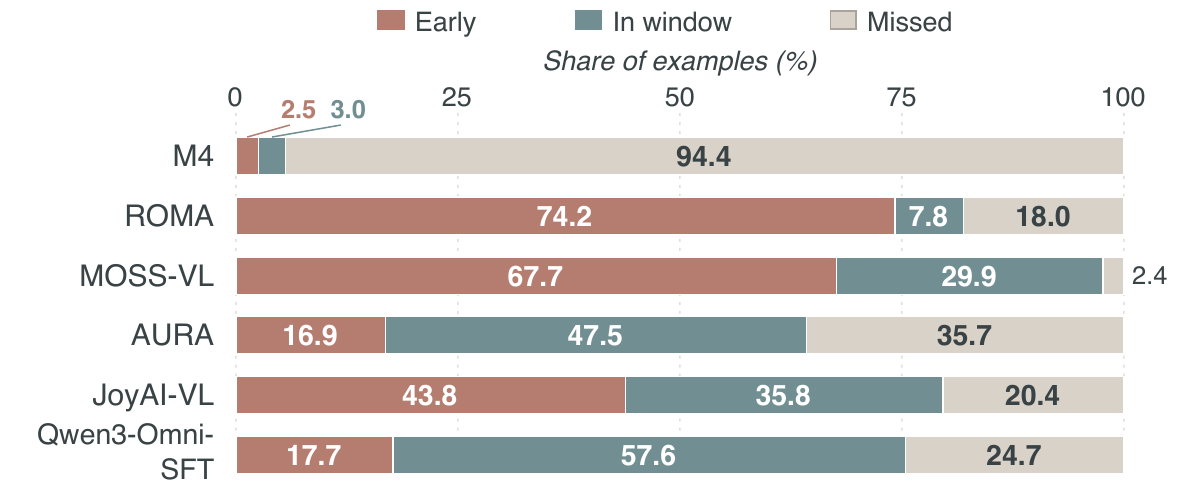}
\caption{Window verdicts for the 628 OR, WR, AA, and SS examples under the audio-disabled primary comparison.}
\label{fig:window}
\end{figure}

\begin{table*}[t]
\caption{Main results (\%). ES reports response and silence rates with their geometric mean. AVG averages the six subtask scores, and audio-disabled rows define the primary comparison.}
\label{tab:main}
\vspace{2pt}
\centering
\footnotesize
\setlength{\tabcolsep}{2.8pt}
\begin{tabular*}{0.95\textwidth}{@{\hspace{8pt}\extracolsep{\fill}}lcccccccccc@{\hspace{8pt}}}
\toprule
\multirow{2}{*}{Model} & \multirow{2}{*}{Audio} & \multicolumn{3}{c}{ES} & \multirow{2}{*}{OR} & \multirow{2}{*}{WR} & \multirow{2}{*}{AA} & \multirow{2}{*}{SS} & \multirow{2}{*}{DC} & \multirow{2}{*}{AVG} \\
\cmidrule(lr){3-5}
 &  & spk & sil & score &  &  &  &  &  &  \\
\midrule
Human Agent & $\checkmark$ & 81.4 & 100.0 & 90.3 & 96.3 & 87.5 & 98.9 & 85.9 & 83.6 & 90.42 \\
\midrule
M4 & $\times$ & 3.7 & 95.5 & 18.8 & 4.6 & 1.3 & 0.8 & 6.5 & 0.0 & 5.30 \\
\multirow{2}{*}{ROMA}
 & $\checkmark$ & 32.9 & 61.8 & 45.1 & 19.3 & 6.3 & 9.2 & 2.2 & 3.0 & 14.16 \\
 & $\times$ & 49.5 & 48.0 & 48.7 & 13.3 & 11.3 & 3.4 & 3.3 & 3.0 & 13.82 \\
MOSS-VL-Realtime & $\times$ & \textbf{82.9} & 50.4 & \textbf{64.6} & 38.1 & 35.0 & 24.8 & 19.6 & 7.5 & 31.60 \\
AURA & $\times$ & 34.3 & \textbf{99.2} & 58.3 & \textbf{62.8} & \textbf{60.0} & 37.8 & 25.0 & 0.0 & 40.70 \\
JoyAI-VL-Interaction & $\times$ & 47.2 & 78.9 & 61.0 & 52.3 & 47.5 & 18.9 & 30.4 & 35.8 & 41.00 \\
\multirow{2}{*}{Qwen3-Omni-SFT}
 & $\checkmark$ & 53.7 & 71.1 & 61.8 & 45.9 & 38.8 & 43.7 & 20.7 & 32.8 & 40.60 \\
 & $\times$ & 41.7 & 84.6 & 59.4 & 52.3 & 43.8 & \textbf{73.1} & \textbf{42.4} & \textbf{37.3} & \textbf{51.37} \\
\bottomrule
\end{tabular*}
\end{table*}

Under the audio-disabled primary setting, Qwen3-Omni-SFT achieves the highest AVG of 51.37 (Table~\ref{tab:main}), leading AA, SS, and DC. MOSS-VL-Realtime leads ES, while AURA leads OR and WR, indicating that no system dominates all evaluated behaviors.

\textbf{Human performance sets a non-trivial ceiling.}
The Human Agent achieves 92.99\% AVG, validating task clarity, yet scores lower on SS (85.9\%) and DC (83.6\%), reflecting the inherent cognitive load of sustained vigilance and counting. The best model trails by over 40 points, with deficits concentrated in these same temporally demanding subtasks, confirming that current systems lack human-like temporal decision policies.

\textbf{Temporal errors are dominated by premature responses.}
The aggregate scores conceal distinct temporal error profiles. In Fig.~\ref{fig:window}, early responses outnumber missed responses for ROMA, MOSS-VL-Realtime, and JoyAI-VL-Interaction by factors of 2.1--28.2. Qwen3-Omni-SFT instead yields 57.6\% in-window, 17.7\% early, and 24.7\% missed responses. Timestamp-selected queries omit responses before the target window and may therefore obscure this error mode and alter model rankings.

\textbf{Audio significantly impacts the response timing of omni models in distinct ways.}
Audio changes the response policies of Qwen3-Omni-SFT and ROMA in opposite directions. For Qwen3-Omni-SFT, enabling audio increases the response rate from 6.5\% to 10.2\% and early responses from 17.7\% to 48.4\%, while its ES score remains comparable (59.4\% vs. 61.8\%). In contrast, removing audio from ROMA increases the response rate from 11.2\% to 13.1\%, reduces the mean delay from 0.56 to 0.33~s, and increases early responses from 66.2\% to 74.2\%. Its ES response component rises from 32.9\% to 49.5\%, while the silence component decreases from 61.8\% to 48.0\%. Overall, these results show that audio can shift response timing in model-dependent ways.

{\clubpenalty=10000
\textbf{Geometric aggregation penalizes imbalanced response policies.}
ES further demonstrates the value of jointly evaluating response and silence to prevent degenerate strategies from achieving deceptively moderate scores. M4's rates of 3.7\% and 95.5\% yield an arithmetic mean of 49.6, effectively masking its inability to trigger timely alerts, but a geometric mean of 18.8 properly exposes this severe imbalance. ROMA's audio-disabled rates of 49.5\% and 48.0\% instead yield a geometric mean of 48.7, matching its balanced trade-off.
\par}

\section{Limitations}

This benchmark evaluates end-to-end system behavior, inherently coupling response policy with perceptual capability. Comparisons are further shaped by heterogeneity in model modalities and silence mechanisms, as well as fixed evaluation parameters that define the current temporal resolution and historical scope. Delay statistics are conditioned on successful in-window responses, and while the six evaluated systems reveal consistent task-specific challenges, broader generalization of error patterns remains an open question for future investigation. Future work should incorporate uncertainty quantification and systematically assess sensitivity to context length, response-window definitions, and model-specific decision thresholds.

\section{Conclusion}
\label{sec:conclusion}

We present LiveProBench, a benchmark for evaluating response timing in streaming multimodal models under standing requests. Unlike discrete timestamp-based protocols, LiveProBench continuously evaluates both response initiation and intentional silence throughout the full stream, requiring models to decide when to speak without an explicit response cue. Across six systems, premature responses outnumber missed responses for three systems in the audio-disabled comparison, revealing a prevalent tendency to respond before sufficient evidence has accumulated. These findings highlight that effective streaming multimodal interaction requires not only recognizing events, but also making timely decisions about when to speak and when to remain silent. LiveProBench provides a time-resolved framework for measuring this response policy and offers a step toward evaluating whether streaming multimodal models can interact with continuous streams in a more human-like manner.

\section{Compliance with Ethical Standards}

This work reports computational evaluation on existing video recordings and does not involve intervention with human participants or animals. No institutional ethics approval was required.

\section{Acknowledgments}

The authors declare no conflicts of interest. Generative AI tools assisted with drafting, language editing, and figure preparation; the authors reviewed and approved all technical content, analyses, and conclusions.

\bibliographystyle{IEEEbib}
\bibliography{streaming_refs}

\end{document}